\documentclass{article}
\usepackage{ijcai23}

\usepackage{times}
\usepackage{soul}
\usepackage{url}
\usepackage[hidelinks]{hyperref}
\usepackage[utf8]{inputenc}
\usepackage[small]{caption}
\usepackage{graphicx}
\usepackage{amsmath}
\usepackage{amsthm}
\usepackage{booktabs}
\usepackage{algorithm}
\usepackage{algorithmic}
\usepackage[switch]{lineno}

\usepackage{subcaption}
\usepackage{amssymb}
\usepackage{array,booktabs}
\newcommand {\otoprule}{\midrule [\heavyrulewidth]}
\newcolumntype {+}{ >{\global\let\currentrowstyle\relax}}
\newcolumntype {^}{ >{\currentrowstyle }}
 \newcommand {\rowstyle}[1]{\gdef\currentrowstyle{#1} %
 #1\ignorespaces
 }
\newcommand{\tabhead}{\rowstyle{\bfseries}}
\usepackage{makecell}
\usepackage[inline]{enumitem}

\title{Is Last Layer Re-Training Truly Sufficient for Robustness to Spurious Correlations?}

\author{
Phuong Quynh Le$^1$
\and
Jörg Schlötterer$^{1,2,3}$\And
Christin Seifert$^{1,2}$\\
\affiliations
$^1$University of Duisburg-Essen, Germany\\
$^2$University of Marburg, Germany\\
$^3$University of Mannheim, Germany
\emails
phuongquynh.le@uni-due.de,
\{joerg.schloetterer, christin.seifert\}@uni-marburg.de
}

\begin{document}

\maketitle

\begin{abstract}
Models trained with empirical risk minimization (ERM) are known to learn to rely on spurious features, i.e., their prediction is based on undesired auxiliary features which are strongly correlated with class labels but lack causal reasoning. 
This behavior particularly degrades accuracy in groups of samples of the correlated class that are missing the spurious feature or samples of the opposite class but with the spurious feature present.
The recently proposed Deep Feature Reweighting (DFR) method improves accuracy of these worst groups.
Based on the main argument that ERM mods can learn core features sufficiently well, DFR only needs to retrain the last layer of the classification model with a small group-balanced data set. 
In this work, we examine the applicability of DFR to realistic data in the medical domain. 
Furthermore, we investigate the reasoning behind the effectiveness of last-layer retraining and show that even though DFR has the potential to improve the accuracy of the worst group, it remains susceptible to spurious correlations. 
\end{abstract}

\section{Introduction}
Machine Learning models are known to be susceptible to learning spurious correlations~\cite{Geirhos2020_nature_shortcut-learing}. A spurious attribute is a characteristic that is present in the majority of training data for a specific class, but has no causal relationship to the class label. A standard empirical risk minimization (ERM) model which learns to minimize the average loss may fail to obtain high \textit{worst-group} accuracy. 
For example, image classifiers that rely on background to predict the class of an object in the foreground may fail when objects are placed onto other backgrounds~\cite{why-should-i-trust-you,zhou2021examining}; gender prediction models that correlate male celebrities with dark hair show surprisingly low accuracy in the group of males with blond hair~\cite{sagawa2019distributionally}. With increasing uptake of machine learning in high-risk domains such as healthcare~\cite{davenport2019potential,healthcare-review}, e.g., for cancer diagnosis~\cite{nassif2022breast,lung-cancer,dildar2021skin}, spurious correlations become particularly concerning. In fact, \cite{winkler2019association} found that a model with nearly 100\% accuracy in melanoma classification from dermoscopic images was sensitive to the presence of surgical markers, which are frequently found in suspicious lesions.

In order to improve the worst-group accuracy, \cite{sagawa2019distributionally} combined strong regularization with distributionally robust optimization (DRO), which is focused on minimizing worst-group loss instead of average loss.
However, this approach requires knowledge of spurious attributes prior to training and access to all group labels for robust loss calculation, which can be expensive and time-consuming to obtain.
To address these limitations, \cite{sohoni2020no} automatically infer groups by clustering the feature space of the deep neural network before applying the DRO loss function. 
Similarly, \cite{creager2021environment} focus on inferring appropriate groups before training a robust model.
Simpler, yet effective approaches involve training two ERM models. The first model is used to identify hard-to-learn training data points by detecting mis-classified instances. The second ERM model is then trained using these hard data points by either up-weighting this group~\cite{nam2020learning,liu2021just} or fine-tuning with respect to this group exclusively~\cite{yaghoobzadeh2019increasing}. 
All these methods significantly improve worst-group accuracy without the need for explicit group labels, but perform worse than methods that rely on explicit group labels.

While the techniques mentioned above fall into two categories of extremes, either requiring annotation of all training data with group labels or not requiring group label annotation at all, BARACK~\cite{sohoni2021barack} and DFR~\cite{kirichenko2023last} have recently introduced a third category that requires only a limited number of group label data. 
In particular DFR shows great potential as it minimizes the retraining effort by adjusting only the weights of the final layer while achieving similar or superior performance compared to DRO, motivating our choice of analysing DFR in this paper.

The main contributions of this paper are as follows. First, we reproduce DFR using a benchmark dataset and a realistic dataset and demonstrate its efficacy in improving  worst-group accuracy. Second, we conduct a detailed analysis of the model's success using visualisation techniques. Our analysis reveals that, despite the model's simplicity and ease of application, it fails to eliminate all spurious correlations, highlighting its limitations.

\begin{table}[t]
\begin{tabular}{+l^c^c}
\toprule \tabhead
        & Waterbirds     & ISIC Skin  \\\otoprule
\( \mathcal{Y} \)       & waterbird / landbird  & malignant / benign \\
\( \mathcal{S} \)       & background (water, land) & colored patch \\
Train.                   & 1,102 / 3,693       &  1,914 \& 14,445 \\
                        & {\small(\%water-bg) 95\% / 5\%}   & {\small(\%patch) 0\% / 46\%} \\
Valid.                  & 372 / 827    &  120 / 120 \\
                        & {\small(\%water-bg) 50\% / 50\%}   & {\small (\%patch) 50\% / 50\%} \\
Test                    & 1,622 / 4,172       &  958 / 3316 \\
                        & {\small (\%water-bg) 50\% / 50\%}   & {\small (\%patch) 50\% / 50\%} \\
\bottomrule
\end{tabular}
\caption{Overview of data sets. \( \mathcal{Y} \): classes for classification task.
\( \mathcal{S} \): type of spurious correlation in the data set.
Train., Valid., Test: number of instances per class \textit{waterbird / landbird} and \textit{malignant / benign}, and the percentage of images with the spurious correlation feature present (e.g.,  in Waterbirds, 1,102 images of class waterbird are present, 95\% of which on a water background). In validation and test sets, spurious correlation features are balanced (50\%).
}
\label{tab:dataset}
\end{table}

\section{Methodology}

Deep Feature Reweighting (DFR)~\cite{kirichenko2023last} is a two-stage method resolving spurious correlations in classification tasks. 
With the main argument that empirical risk minimization (ERM) is capable of acquiring feature representations well and independently, the authors state, that it is sufficient to adjust the weights of the final fully connected layer, which defines the importance of each feature on a particular class. 
First, a standard ERM model is trained including \( f_{enc}: X \mapsto \mathbb{R}^d \) the feature representation encoder and \( f_{cls}: \mathbb{R}^d \mapsto \mathbb{R}^C \) the fully connected last layer.
Second, to mitigate spurious correlations, only the final layer $f_{cls}$ is re-trained using a logistic regression model.
The input to the logistic regression model is the feature representation as learned by ERM $z=f_{enc}(\mathrm{x})=[z_1, z_2, ..., z_d]$.
For mitigating spurious correlations, the re-training phase requires a group-balanced subset of data, i.e., a small dataset unaffected by spurious features.
The group-balanced dataset is obtained by selecting an equal number of samples from each group of \( \mathcal{G} \) in the validation set.

We use a pre-trained Resnet50~\cite{resnet} as base model and adopt hyper-parameters from~\cite{kirichenko2023last}. 
We replace the final softmax layer of the pre-trained Resnet50 by a single sigmoid neuron to adapt to the binary classification task and to have comparable last layer weights between the ERM base model and logistic regression which weight parameters defined as a vector \( [w_1, w_2, \ldots w_d] \). 

We report accuracy per group and overall for model evaluation.
We highlight \textit{worst-group accuracy}, used to evaluate robustness to spurious correlations.
From here to the rest of this paper, we define the term \textit{worst-group accuracy} as the group with the lowest test accuracy in the ERM model. 

For the qualitative analysis, we use class activation maps (CAM)~\cite{zhou2016cam} to localize important regions in the original input image $\mathrm{x}$, both for ERM and DFR models.
On image level, we take the whole feature representation and last layer weights into account.
For a detailed analysis of individual neurons, we visualize the activation map of a single neuron out of the $d$ neurons in the feature representation layer.

\section{Data Sets} \label{sec:dataset}
\paragraph{Waterbirds.} Waterbirds was introduced by \cite{Sagawa2019waterbird} as a standard spurious correlation benchmark. This dataset comprises the CUB dataset~\cite{wah2011caltech} and the Places dataset~\cite{place-dataset}, with the bird segmentation from CUB pasted onto the background of the Places dataset. The birds labeled as either waterbirds or landbirds are placed on backgrounds of either water or land. This task aims to classify waterbirds and landbirds: \( \mathcal{Y} \) = \{waterbird, landbird\}.Placing 95\% waterbirds on water background and 95\% landbirds on land background results in spurious correlations in the default training dataset since model tends to rely on background to predict instead of actual bird characteristics. From there, the spurious feature is defined as \( \mathcal{S} \) = \{water background, land background\}. The distribution of water and land backgrounds among the two bird classes is even for validation and test sets which means the spurious correlation is 50\%. The set of groups $\mathcal{G}$ is defined as the set of all possible pairs of class label and spurious attribute, i.e.,  \( \mathcal{G}  = \mathcal{Y} \times \mathcal{S}\).
\begin{table}[t]
\centering
\begin{tabular}{+l^c^c}
\toprule \tabhead
 & ERM & DFR\\
\midrule
\multicolumn{3}{c}{\textbf{ISIC Skin}} \\
\midrule
Benign w/o patch & $94.29_{\pm 0.82}$ & $77.72_{\pm 2.39}$ \\
Benign with patch & $100_{\pm 0.00}$ & $95.77_{\pm 1.44}$ \\
\textbf{Malignant w/o patch}&$\mathbf{64.38}_{\pm 1.44}$&$\mathbf{85.84}_{\pm 1.76}$ \\
Malignant with patch & $65.00_{\pm 5.20}$ & $98.61_{\pm 0.69}$ \\
Average & $90.12_{\pm 0.79}$ & $87.43_{\pm 1.34}$ \\
\midrule
\multicolumn{3}{c}{\textbf{Waterbirds}} \\
\midrule
Landbird on land & $99.56_{\pm 0.04}$ & $95.56_{\pm 0.91}$ \\
Landbird on water & $86.49_{\pm 1.58}$ & $91.33_{\pm 0.89}$ \\
\textbf{Waterbird on land} & $\mathbf{72.86}_{\pm 1.74}$ & $\mathbf{92.55}_{\pm 0.44}$ \\
Waterbird on water & $96.53_{\pm 0.74}$& $95.07_{\pm 0.89}$ \\
Average & $91.17_{\pm 0.52}$ & $93.53_{\pm 0.62}$ \\
\bottomrule
\end{tabular}
\caption{Group and average accuracy (mean and standard deviation over five seeds) of ERM and DFR models on dataset ISIC Skin and Waterbirds. Worst-group in \textbf{bold}.}
\label{tab:llr_result}
\end{table}
\paragraph{ISIC Skin.} ISIC Skin is a real-world medical dataset provided by the International Skin Imaging Collaboration ISIC~\cite{isic2019}. The objective is to distinguish benign (non-cancerous) cases from malignant (cancerous) cases. Using the source code of~\cite{rieger2020}, we retrieved 20,394 images for the two classes in the two categories benign (17,881) and malignant (2,513) from the ISIC Archive\footnote{\url{https://www.isic-archive.com/}}. Nearly half of the benign cases (8349) have colored patches attached to patients' skin, whereas no malignant case contains such patches.
Since colored patches appear only in benign cases and are considered to be easier to learn~\cite{diagnostics12010040}, we define the target label \( \mathcal{Y} \) = \{benign, malignant\} and spurious attribute \( \mathcal{S} \) = \{with patch, without patch\}. 
To construct the missing group (malignant with patch), we use the same method as previous  work to add colored patch segments to 60 and 479 malignant images in the validation test set, respectively.
More details of datasets are in Table \ref{tab:dataset}.

\section{Results}
In this section, we analyse the effects of Deep Feature Reweighting (DFR) through
\begin{enumerate*}[label=(\roman*), itemjoin={{, }}, itemjoin*={{, and }}]
  \item a quantitative analysis by group accuracy
  \item a qualitative analysis on neuron selection by visualising last layer weights and activation maps.
  \end{enumerate*}
\subsection{DFR Performance}
Table~\ref{tab:llr_result} shows the accuracy per group and micro-averaged over groups of ERM and DFR for ISIC Skin and Waterbirds. 
We observe that DFR significantly improves the worst-group accuracy, i.e, waterbird on land from 72.86\% to 92.55\% and malignant w/o patch from 64.38\% to 85.84\%. 
In the Waterbirds dataset, we further observe a performance gain for landbird on water, the other pair of (spurious feature, uncorrelated class). At the same time, performance in the correlated classes decreases only slightly, yielding an improvement in overall accuracy.
In the ISIC Skin dataset, while the worst group accuracy improves by a large margin, we simultaneously observe a significant drop in the accuracy of the benign w/o patch group, resulting in a slight decrease in overall accuracy.
Although the improvement of worst-group accuracy is typically regarded as a measure of the efficacy of a method that is robust to spurious correlations, we question whether all unwanted features are eliminated from the reasoning of predictions.
To understand the extent of performance differences, we analyse and interpret how the weight distribution changes after retraining w.r.t. the question of whether \textit{all spurious features are removed by retraining only the last layer?}

\subsection{Analysis of Last Layer Weights}
The last layer plays a crucial role in determining how the feature representations are combined and transformed to produce predictions for target classes. As a result, modifying the weights of the last layer can significantly alter the distribution of important features associated with each class~\cite{kirichenko2023last}.

\begin{figure}
     \centering
     \begin{subfigure}[b]{0.49\columnwidth}
         \centering
         \includegraphics[width=\linewidth]{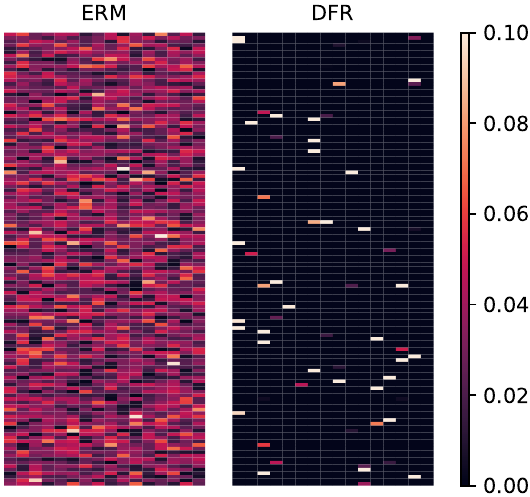}
         \caption{\footnotesize ISIC Skin}
         \label{fig:isic-weight}
     \end{subfigure}
     \hfill
     \begin{subfigure}[b]{0.49\columnwidth}
         \centering
         \includegraphics[width=\linewidth]{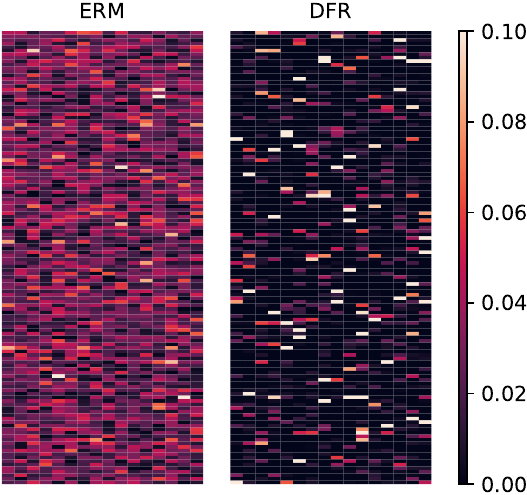}
         \caption{\footnotesize Waterbirds}
         \label{fig:wb-weight}
     \end{subfigure}
        \caption{Last layer weights heatmap of ERM and DFR models (the weight vector is reshaped to two dimensions in visualisation).}
        \label{fig:heatmap}
\end{figure}
To understand the effect of DFR, we visualize the weights of the last layer before and after the retraining step.
Figure~\ref{fig:heatmap} shows the weights of the last layer before and after DFR for a random run for each dataset. We observe an unexpectedly high percentage of connections with zero weight in DFR (96\% for ISIC Skin and 65\% for Waterbirds, on average across five runs for each dataset), which implies that DFR identifies a significant number of irrelevant features in the feature representation and eliminates their impact for classification. 
In the context of pruning techniques, it is noteworthy that pruning up to 60-70\% of model parameters was reported to keep overall accuracy almost unchanged~\cite{shiweiliu-pruning,zhao2019variational}, which is comparable to the amount of pruned last layer weights for Waterbirds and significantly less than the amount of pruning in ISIC Skin.\footnote{The mentioned techniques refer to pruning the whole model parameters (instead of a single layer only) and are not intended to address spurious correlations.}

\subsection{Qualitative Analyses} \label{visulaisation}
\begin{figure}[t]
    \centering
 \includegraphics[width=0.7\columnwidth]{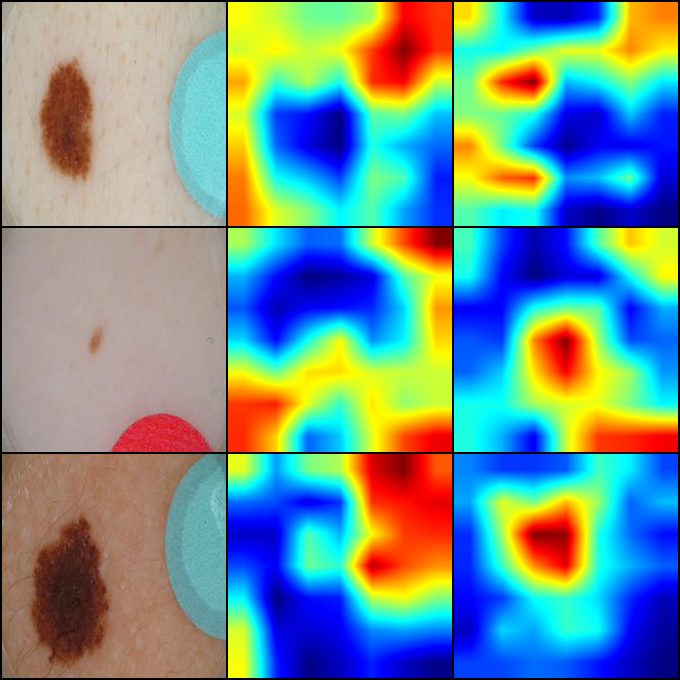}
 \caption{CAM visualisation for three exemplary test images.  Showing original image (left), CAM before retraining (center), and CAM after retraining. Red indicates high activation, blue low activation.}
 \label{fig:cam}
\end{figure}

\paragraph{Image-level.}
Figure \ref{fig:cam} shows exemplary CAM visualisation from ISIC Skin. Each row corresponds to one test image: original image (left), activation map corresponding to ERM model (middle) and activation map after retraining with DFR (right). To evaluate the sensitivity of ERM and DFR to the spurious attribute (colored patch), we only select instances from the benign with patch group.
Although DFR has shown success in redirecting attention away from unwanted features, i.e., colored patches, and towards core features, i.e., lesions (third row), there are instances in which DFR appears to be influenced by spurious correlations. More specifically, as shown in the first and second row, DFR puts on stronger focus (higher activation) on core features, i.e., the lesion. Still, it takes the spurious component into consideration (e.g., high activation at the bottom right corner of third image in row two), although to a lesser overall extent than core features.

\begin{figure*}
\centering
\renewcommand{\arraystretch}{1.5}
\begin{tabular}{+>{\centering\arraybackslash}m{1.2cm}^>{\centering\arraybackslash}m{3.4cm}^>{\centering\arraybackslash}m{3.4cm}^>{\centering\arraybackslash}m{3.4cm}^>{\centering\arraybackslash}m{3.4cm}}

	 & \textbf{Example Images} & \makecell{\textbf{Spurious only}\\(Neuron 2014)} & \makecell{\textbf{Core Only} \\ (Neuron 1986)} & \makecell{\textbf{Core + Spurious} \\ (Neuron 222)} \\
 Benign w/o patch 
 	& \includegraphics[width=3.4cm]{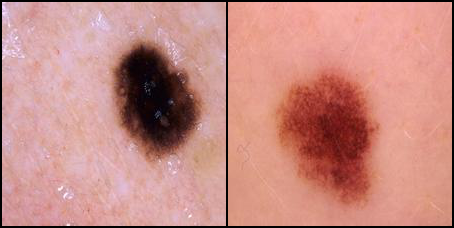}
	& \includegraphics[width=3.4cm]{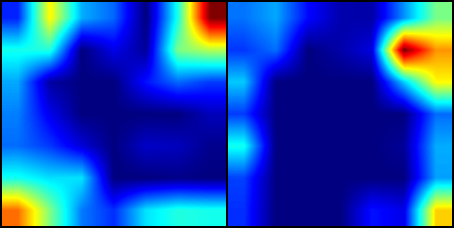}
    & \includegraphics[width=3.4cm]{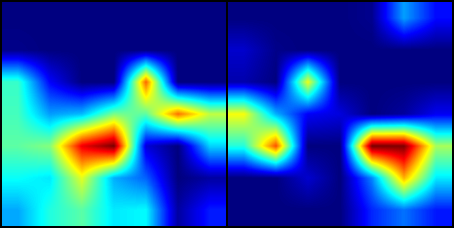}
    & \includegraphics[width=3.4cm]{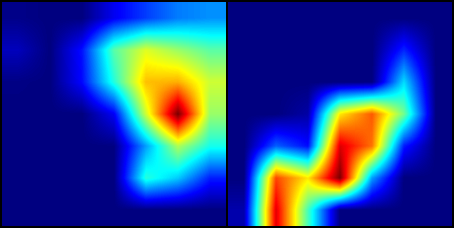} \\
 Benign w patch 
	& \includegraphics[width=3.4cm]{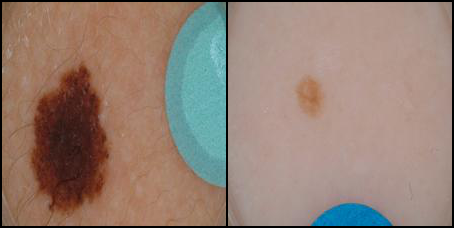} 
    & \includegraphics[width=3.4cm]{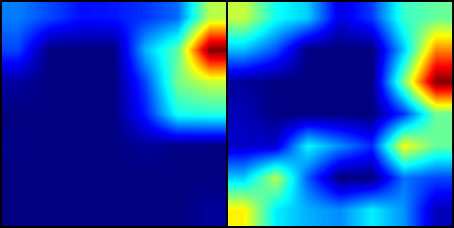}
    & \includegraphics[width=3.4cm]{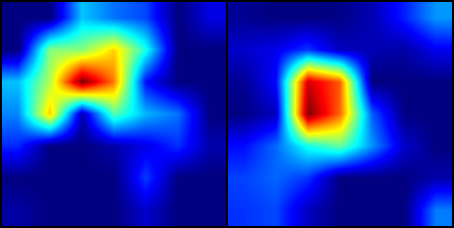}
    & \includegraphics[width=3.4cm]{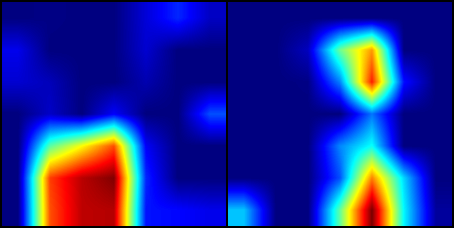}\\
Malignant w/o patch
  	& \includegraphics[width=3.4cm]{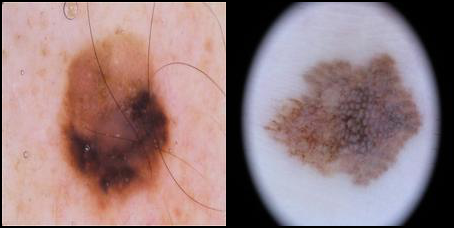}   
  	& \includegraphics[width=3.4cm]{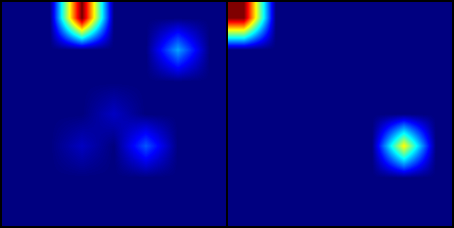}
  	& \includegraphics[width=3.4cm]{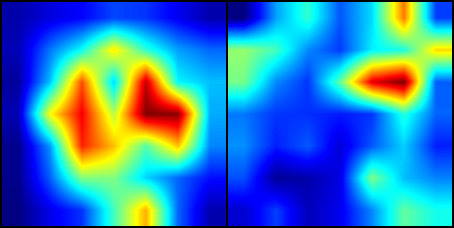}
  	& \includegraphics[width=3.4cm]{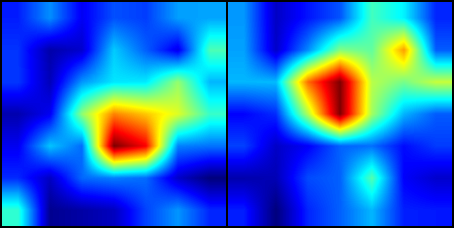}\\
Malignant w patch  
	& \includegraphics[width=3.4cm]{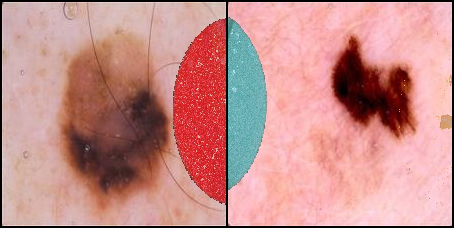}   
	& \includegraphics[width=3.4cm]{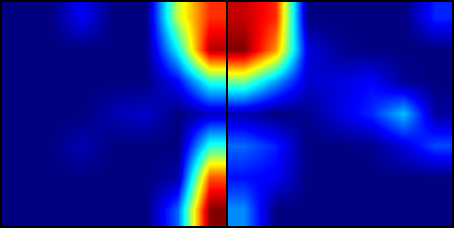}
	&\includegraphics[width=3.4cm]{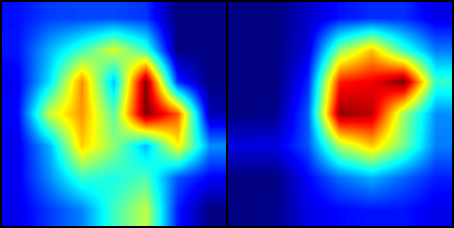}
    & \includegraphics[width=3.4cm]{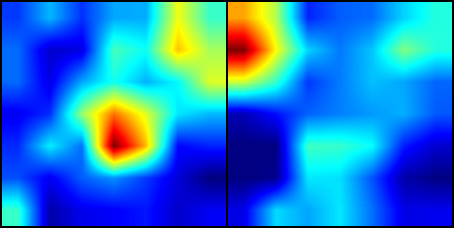}\\
\end{tabular}
\caption{Neuron-level visualisations. Showing two example for each group and activations for three exemplary neurons from the last layer, one neuron that encodes spurious features, one neuron that encodes core features and one neuron that encodes both, core and spurious features.}
\label{tab:cam-neurons}
\end{figure*}

The last layer becomes significantly more compact (i.e., many weights are pruned) after applying DFR, which implies that DFR effectively removes the major impact of spurious correlations as expected. However, the weighted activation maps of the input do not necessarily support the removal of all spurious correlations. Therefore, a thorough examination of individual neurons that were (not) selected as important by DFR is required.

\paragraph{Neuron-level.}
We found that among the $d$ neurons of feature representation, there are three types of neurons which are spurious only, core only and a mixture (core + spurious).

Spurious-only neurons are those that primarily focus on spurious features (the color patch) or unrelated features (which could be patients' skin area or image boundary localisation). The column \textbf{Spurious only} in Figure~\ref{tab:cam-neurons} provides an example of a targeted neuron that always looks at patch localisation or edges but focuses very little on the lesion area. This type of neuron is identified by examining neurons that have been turned off by DFR, i.e., those with zero weight after the retraining phase.

Among retained neurons of DFR, we investigate those that have a high connected last layer weight magnitude, i.e., a greater impact on class prediction. Evidently, it is possible to find core-only neurons, that focus on the lesion only, even in the presence of patches; an example is provided in column \textbf{Core only} Figure \ref{tab:cam-neurons}.
The above spurious-only and core-only types of neurons suggest that, first, ERM is capable of learning core and spurious features separately to some extent. Second, DFR is capable of identifying these easy-case neurons.

However, there are neurons with complex responses, which DFR is incapable of distinguishing. It appears that DFR partially selects this type of neurons which can respond to both, lesion and color patches. Specifically, this neuron type behaves appropriately when only a lesion is present (focusing solely on the center of the lesion), but promptly changes focus when a patch is introduced (highlights the patch rather than the core lesion). Examples are shown in column \textbf{Core + Spurious} of Figure \ref{tab:cam-neurons}. This finding suggests that even though DFR improves the worst-group accuracy by a large margin, its reasoning may still be flawed.

\section{Conclusion}
In this work, we reproduced and analysed the effectiveness of Deep Feature Reweighting (DFR), a two-step approach robust to spurious correlations on a realistic dataset from the medical domain. Our findings show that the model is effective in improving the worst-group accuracy, but it does not completely eliminate all spurious correlations. These findings on ISIC Skin cannot be extrapolated to general cases and all other datasets, but they do suggest additional research on the topic and future prospects, particularly in the medical domain, where truthfulness is of utmost importance. In this paper, we evaluate core-only neurons based on whether they examine the true lesion region. Other essential features with regard to the cancer diagnosis domain include Asymmetry, Border, Color, Diameter - the ABCD rule~\cite{abcd-rule2021} can also be taken into account. We recommend that future research incorporate professional medical knowledge for deeper comprehension of reasoning.
Moreover, the effectiveness of feature attribution techniques remains debatable~\cite{zhou2022feature}. We would not rule out the possibility that the activation maps themselves are insufficiently accurate; thus, additional XAI approaches are encouraged.

DFR is easy to implement and its requirement for a small number of group label annotation are advantages. The failure to remove all spurious attributes during retraining the last layer, however, suggests that either the method chosen for retraining (logistic regression) is insufficient for the core-only neuron selection, or the representation learned by empirical risk minimization (ERM) itself is severely affected by spurious correlations and lacks a sufficient learning of core features for accurate prediction. Our analysis highlights the need for more advanced methods to address the problem of spurious correlations in machine learning models.

\clearpage

\bibliographystyle{named}
\bibliography{ijcai23}

\end{document}